\documentclass[conference]{IEEEtran}
\usepackage{booktabs}
\usepackage{enumitem}
\usepackage{amsmath}
\usepackage{graphicx}
\usepackage{float}
\usepackage{placeins}
\usepackage{subcaption}
\usepackage{hyperref}
\usepackage{textcomp}
\usepackage{comment}
\usepackage[utf8]{inputenc}
\usepackage[
  style = ieee,    
  sorting = none,      
  maxbibnames = 3,     
  minbibnames = 1      
]{biblatex}
\AtBeginBibliography{\footnotesize}
\AtEveryBibitem{%
\clearfield{language}%
\clearfield{number}%
\clearfield{pages}%
\clearfield{volume}%
\clearfield{issn}%
\clearfield{urldate}%
\clearfield{note}%
  \clearfield{doi}%
  \clearfield{url}%
  \clearfield{isbn}%
  \clearfield{eprint}%
}

\ifCLASSOPTIONcompsoc
\else
\fi
\ifCLASSINFOpdf

\else

\fi

\begin{document}
%
\title{%
  WatchAnxiety: A Transfer Learning Approach for State Anxiety Prediction from Smartwatch Data
  \vspace{-0.25cm}%
  }

\author{
  Md Sabbir Ahmed$^{1}$ , Noah French$^{2}$, Mark Rucker$^{1}$, Zhiyuan Wang$^{1}$, Taylor Myers-Brower$^{2}$,\\ 
  Kaitlyn Petz$^{2}$, Mehdi Boukhechba$^{3}$, Bethany A. Teachman$^{2}$, Laura E. Barnes$^{1}$\\
  \vspace{-5.5pt}
  \\
$^{1}$Dept. of Systems \& Information Engineering, $^{2}$Dept. of Psychology,\\
University of Virginia, USA\\
$^{3}$Johnson \& Johnson Innovative Medicine, USA\\
\{msabbir, njf5cu, mr2an, vmf9pr, cdf3zf, kdp8y, bteachman, lb3dp\}@virginia.edu; mboukhec@its.jnj.com
\vspace{-0.5em}
  }

\maketitle

\begin{abstract}
Social anxiety is a common mental health condition linked to significant challenges in academic, social, and occupational functioning. A core feature is elevated momentary (state) anxiety in social situations, yet little prior work has measured or predicted fluctuations in this anxiety throughout the day. Capturing these intra-day dynamics is critical for designing real-time, personalized interventions such as Just-In-Time Adaptive Interventions (JITAIs). To address this gap, we conducted a study with socially anxious college students (N=91; 72 after exclusions) using our custom smartwatch-based system over an average of 9.03 days (SD = 2.95). Participants received seven ecological momentary assessments (EMAs) per day to report state anxiety. We developed a base model on over 10,000 days of external heart rate data, transferred its representations to our dataset, and fine-tuned it to generate probabilistic predictions. These were combined with trait-level measures in a meta-learner. Our pipeline achieved 60.4\% balanced accuracy in state anxiety detection in our dataset. To evaluate generalizability, we applied the training approach to a separate hold-out set from the TILES-18 dataset—the same dataset used for pretraining. On 10,095 once-daily EMAs, our method achieved 59.1\% balanced accuracy, outperforming prior work by at least 7\%.

\end{abstract}

\IEEEpeerreviewmaketitle

\vspace{-3pt}
\section{Introduction}
\vspace{-3pt}
Social anxiety, or anxiety tied to social situations in which one may be evaluated negatively, is a prevalent mental health problem. An estimated 12.1\% of individuals in the U.S. meet the criteria for social anxiety disorder at some point in their life \cite{kessler_twelvemonth_2012}. Social anxiety often limits individuals’ lives and is associated with avoiding potentially meaningful careers that require social interactions, avoiding romantic relationships, and delaying starting families \cite{caspi_moving_1988}. Existing research shows that helping people respond to state anxiety in more effective ways (e.g., by challenging anxious thinking and approaching rather than avoiding feared situations) can reduce overall levels of social anxiety \cite{KINDRED2022102640}. However, much of the existing research using passive sensing to detect anxiety has focused on predicting between-person differences in anxiety levels—most commonly trait anxiety, a stable and enduring tendency to experience anxiety across time and situations (e.g., \cite{boukhechba_predicting_2018})—or general anxiety symptoms \cite{sahu_wearable_2024}. While predicting trait anxiety through passive sensing can be useful for early identification of mental health conditions, advancing toward the detection of within-person fluctuations in anxiety (i.e., state anxiety) is essential for enabling real-time, adaptive interventions that address anxiety in the moment.


 Past studies have used passive sensing to predict within-person anxiety level at the daily timescale (e.g., \cite{rashid_predicting_2020}), but only a handful of studies (e.g., \cite{ Larrazabal_Wang_Rucker_Toner_Boukhechba_Teachman_Barnes_2025}) have attempted to estimate within-person fluctuations in anxiety measured at the timescale of hours or minutes, with most such research conducted in controlled laboratory settings. Only one study to our knowledge has attempted to predict within-person fluctuations in anxiety measured multiple times per day outside of a controlled lab setting \cite{jacobson_digital_2022}. However, in \cite{jacobson_digital_2022}, the authors used $R^{2}$ as the evaluation metric, which does not directly reflect predictive accuracy, leaving the model’s effectiveness in identifying moments of state anxiety unclear. Moreover, their models relied on smartphone sensor data, which may be less effective for detecting momentary anxiety, as smartphones are not always carried as frequently as smartwatches \cite{Shahmohammadi2017-hm}.

 For our study, we developed WatchAnxiety, a smartwatch-based system that advances wearable computing and pervasive health by using transfer learning to predict state anxiety. Validated on 2,742 real-world EMA responses, the model achieved 60.4\% balanced accuracy and F1 score. To assess generalizability, we then applied our meta-learning approach to an independent dataset of 10,095 state-anxiety EMAs—bringing the total labeled samples to over 12,000, the largest evaluation to date. This scale is noteworthy given that wearable mental-health research is often hampered by limited labeled data, which can impede robust validation and real-world deployment.


\begin{figure*}[htbp]
    \centering
    \includegraphics[width=1\linewidth]{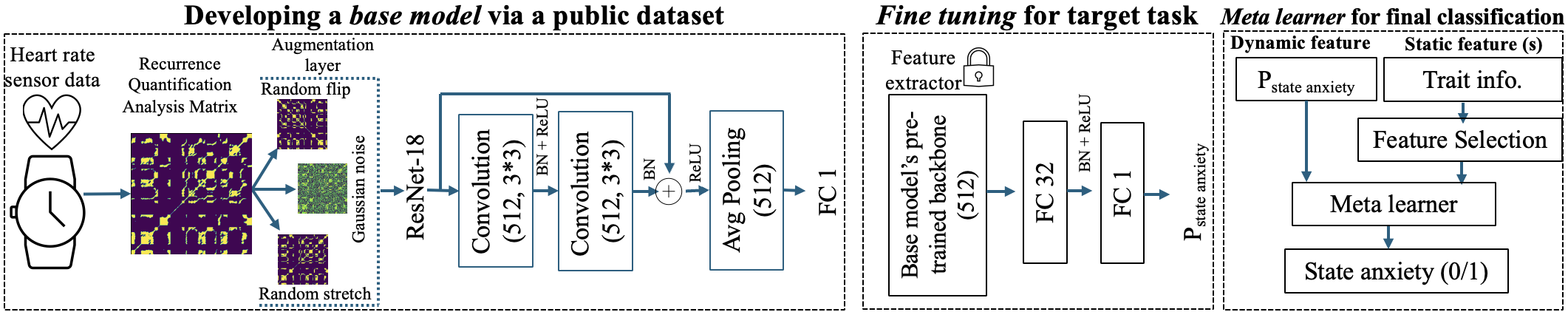}
    \caption{WatchAnxiety system for identifying state anxiety. FC: Fully Connected, BN: Batch Normalization.}
    \label{fig:overall_architecture}
    \vspace{-12pt}
\end{figure*}

\vspace{-3pt}
\section{Methodology}
\vspace{-3pt}
\subsection{Dataset Construction}
\subsubsection{System Design}
 We developed a smartwatch system (expected to be compatible with any Wear OS-based smartwatch) for real-time collection of physiological, behavioral, and acoustic data. To protect participants' privacy, data collection is disabled between 12 AM and 8 AM and automatically pauses when the watch is removed. The system operates on a 5-minute duty cycle, capturing data for 1 minute per cycle. Data are uploaded to secure Amazon S3 storage either manually (via button press) or automatically when the watch is charging and connected to Wi-Fi. Although the system supports multiple sensing modalities, this study focuses exclusively on heart rate (HR)—a widely available physiological marker on commercial smartwatches with strong relevance to health monitoring.

\subsubsection{Participants}
All study procedures were approved by IRB of the University of Virginia (UVA). We recruited participants with moderate-to-severe levels of social anxiety, operationalized as a score of at least 34 on the Social Interaction Anxiety Scale (SIAS) following prior research \cite{toner_wearable_2025}. Each participant was provided with either a Samsung Galaxy Watch 5 or 5 Pro pre-installed with our system. Moreover, participants installed the Sensus \cite{Xiong_Huang_Barnes_Gerber_2016} app on their personal smartphone to receive the EMA surveys. It is worthwhile to mention that early participants received an earlier version of our on-watch system with a duty cycle around 10 minutes, while later participants mostly used the improved version with a 5-minute cycle.

We recruited 91 undergraduate students at UVA. Nineteen participants were excluded due to limited data from the initial version of the system, early withdrawal, or missing watch data within the analysis window. The final sample included 72 participants used for model development.

\subsubsection{Baseline Data Collection}
\label{trait_information}
During the initial study visit, participants completed baseline surveys that captured trait-level mental health characteristics: SIAS, Brief Fear of Negative Evaluation (BFNE), Difficulties in Emotion Regulation (DERS), Depression, Anxiety, and Stress Scales-21 (DASS-21), Adult Rejection Sensitivity Questionnaire (A-RSQ), and Cambridge Depersonalization Scale 2-item version (CDS-2). For missing items (57 item-level responses; 0.59\%), we imputed missing values using the mean of the remaining valid responses of the participant on the corresponding scale. For reverse-scored items, reverse coding was applied prior to both imputation and scale score aggregation. Importantly, imputation was performed independently for each participant, using only their own responses, thereby avoiding the use of data from other participants and preventing data leakage.

\subsubsection{Measuring State Anxiety}
\label{target_variable}
Following the initial study visit, participants reported their state anxiety via ecological momentary assessments (EMAs) delivered through the Sensus smartphone app up to seven times daily for 10 days. EMAs were randomly scheduled every two hours between either 8 AM–10 PM or 10 AM–12 AM, based on participant preference. Each EMA asked, “I feel...,” with responses recorded on a slider from 1 (“not at all anxious”) to 10 (“very anxious”). For classification, responses were binarized: a rating of 1 was coded as class 0 (no anxiety) and all values greater than 1 as class 1 (any level of anxiety).

 Among the 72 participants, HR data were available for 650 total participant-days (mean = 9.03 days, SD = 2.95), with a total of 3,663 state anxiety EMA responses (mean = 50.88 EMAs, SD = 15.91). However, the number of EMA responses included for model development varied based on the explored time window. Specifically, at least 50 HR samples were available within the 1-hour, 1.5-hour, and 2-hour windows for 74.94\%, 75.87\%, and 76.41\% of the EMA responses, respectively. Selecting the appropriate window therefore is critical: larger windows improve data availability but increase overlap between EMAs, while shorter windows may better capture transient physiological markers relevant to state anxiety but increase the chance for data missingness within windows. To balance these trade-offs, we set a 50-sample threshold—an empirically supported cutoff, as nearly all sensor probe start times produced at least 50 HR readings within a one-minute period at $\sim$1 Hz sampling. In other words, if at least one probe occurred within the relevant time window prior to an EMA submission, we included that EMA for model development.

\subsection{Model Development}

\subsubsection{Feature Space}
\label{input_feature_space_creation}
 To construct the input feature space, we first estimate R-R intervals (RRI) from HR using the formula $\text{RRI} = \frac{60}{\text{HR}}$ \cite{Prakash_Madanmohan_2005}. We also excluded HR values outside the physiologically plausible range: above the age-adjusted maximum (220 - age) and below 40 bpm, a threshold reflecting the resting HR of very fit individuals. We then estimated the corresponding RRI timestamps using the cumulative sum of the RRI values, consistent with implementations in widely used packages (e.g., NeuroKit2). Using the inferred RRI and the corresponding timestamps, we performed a recurrence quantification analysis (RQA) of HR variability using the NeuroKit2 package \cite{Makowski_NeuroKit2_2021}. To take advantage of pre-trained ResNet-18 in our base model, we adopted an image-based approach by transforming HR into recurrence plots. These plots are based on time-delayed embeddings of physiological signals, revealing dynamic patterns that can be useful for predicting anxiety.

\subsubsection{Transfer Learning Approach}
\label{base_model_development} 
 Transfer learning (TL) has shown promise across diverse prediction tasks and is particularly beneficial in scenarios with limited data \cite{Ebbehoj_Thunbo_Andersen_Glindtvad_Hulman_2022}. Given our relatively small sample size (N = 72), TL is well-suited. To develop the base model for TL, we used the TILES-18 dataset \cite{Mundnich_Booth_L_et_al._2020}, which includes sensor data from 212 hospital workers collected via multiple devices, including Fitbit. Each day, participants responded to a state anxiety EMA "Please select the response that shows how anxious you feel at the moment" on a scale of 1 to 5. We used the same approach (section \ref{target_variable}) as used for our dataset to create the target variable for classification task. After pre-processing and filtering for entries with at least 50 HR samples, a total of 10,278 EMA responses were available for modeling, with 38.35\% labeled as class 1 and 61.65\% as class 0. However, to explore generalization (section \ref{section_generalization}), the dataset was reduced to 10,095 EMAs due to missing aggregated trait scores in TILES-18 for some participants.

For the model, we adopted ResNet-18 \cite{He_2016_CVPR} without its classification head and initialized it with ImageNet-pretrained weights to leverage transferable representations. We added a residual block, a global average pooling layer, and a final output layer with a single neuron to predict the probability of state anxiety. To address class imbalance, we applied class weights and used the weighted binary focal loss as the loss function. The model was trained with an SGD optimizer with a learning rate of $1e-4$ and trained for a maximum of 20 epochs to obtain reasonable weights for the new layers. We then restored the weights corresponding to the lowest validation loss and fine-tuned the entire model using a reduced learning rate of $1e-8$. To prevent overfitting, early stopping was applied if validation loss did not improve for 3 consecutive epochs.

To reduce computational overhead, we employed a leave-five-out cross-validation (LFOCV) strategy, holding all state anxiety responses of five participants for testing in each fold. Of the remaining participants, two were selected for validation (i.e., used for model selection and early stopping) while the rest were used for training. In some cases, a validation participant reported only one class of state anxiety, which could bias the model. To address this and improve generalization, validation participants were chosen, when available, such that the ratio of class 1 to class 0 was within 10\% of that in the training set.

\subsubsection{Model Tuning}
A key difference between the TILES-18 dataset, used for base model development, and our target task lies in the EMA protocol: TILES-18 collected state anxiety once a day, while our study administered EMA seven times a day to capture intraday fluctuations. TILES-18 also relied on Fitbit devices with continuous data collection, while our custom system employed a duty-cycled sampling strategy to support real-time processing and conserve battery life - enabling both data collection and future on-device interventions. Furthermore, the study populations differed: TILES-18 involved hospital shift workers, while our participants were undergraduate students. 

Since the base models were trained using LFOCV on TILES-18 data, multiple models were generated. We selected the one with the highest balanced accuracy in its respective test set. To adapt the model, we removed the top classification layer and used the 512-neuron global average pooling layer as output. We then added a 32-unit fully connected layer, followed by batch normalization and a ReLU activation function. A final dense layer with a single neuron produced the predicted probabilities. This architecture was inspired by the squeeze-and-excitation (SE) block from SENet, where a low-dimensional representation is learned post-global pooling to enhance generalizable feature learning. The weights of the base model were frozen and the new layers were trained using the Nadam optimizer (learning rate = $1e-5$), as SGD yielded suboptimal results in this context. To avoid overfitting or underfitting, we employed a custom callback that restored the best model based on training dynamics. Specifically, we allowed up to a 3\% tolerance between training and validation loss, restoring the weights with the smallest difference if the validation loss dropped below the training loss or exceeded the tolerance. Training, validation, and testing followed the same LFOCV protocol described in Section \ref{base_model_development}.

\subsubsection{Meta-learner Development and Evaluation}
 After generating predicted probabilities from the fine-tuned model, we incorporated trait measures (Section \ref{trait_information}) to train a meta-learner. This approach is practical for real-world deployment, as trait assessments need to be completed only once prior to system use. To retain only relevant characteristics, we applied feature selection based on information gain to identify the most predictive traits. We then trained lightweight classifiers, K-Nearest Neighbors (KNN), Logistic Regression (Logit), and Decision Tree, as meta-learners for classifying state anxiety. The use of lightweight models was motivated by the goal of minimizing overfitting.

 Given that our meta-learner is lightweight and fast to train, we applied leave-one-out cross-validation (LOOCV) at the final evaluation stage. This setup, in which the intermediate stage uses LFOCV and the final stage uses LOOCV, avoids information leakage. In contrast, using LFOCV in the final stage and LOOCV earlier could introduce leakage by allowing a participant seen during training in intermediate stage to possibly reappear in the test sets of the meta-learner. For model evaluation, we report balanced accuracy, precision, recall, F1-score, and specificity. To address class imbalance, both precision and F1-score were computed as weighted metrics.

\vspace{-3pt}
\section{Results and Discussion}
\vspace{-3pt}
 Although we explored 3 meta-learners, Logit consistently performed better; thus, we report results for Logit only. Across the windows evaluated, the performance was relatively similar (Table~\ref{tab:performance_SAPIENS}); however, the 1.5-hour window offered a favorable balance between recall (58.1\%) and balanced accuracy (60.4\%). To assess robustness, we compared it with two baseline models. Baseline 1 is a model based on all trait measures, while baseline 2 is a random classifier with uniform probability across both classes. As shown in Table \ref{tab:performance_SAPIENS}, our meta-learner using a 1.5-hour window outperformed both baselines. Though baseline 1 has a comparable balanced accuracy (60.4\% vs. 57.3\%) with our meta learner, empirically, we found a model based on trait measures predicted always either class 1 or 0 for all days of each participant (section \ref{ablation_study} for details). \\

\begin{table}[htbp]
\vspace{-6pt}
\caption{Performance of baseline models and our meta-learner. BA = Balanced Accuracy, Prec = Precision, Rec = Recall, Spec = Specificity.}
\label{tab:performance_SAPIENS}
\small
\setlength{\tabcolsep}{4pt}  
\begin{tabular}{lccccccc}
\toprule
Model & EMAs & Trait & Prec. & Rec. & Spec. & BA & F1 \\
\midrule
Meta (1h)   & 2703 & 4 & 61.8 & 57.9 & 61.9 & 59.9 & 60.0 \\
Meta (1.5h) & 2742 & 4 & \textbf{62.3} & \textbf{58.1} & 62.7 & \textbf{60.4} & \textbf{60.4} \\
Meta (2h)   & 2765 & 4 & 62.6 & 56.6 & \textbf{64.8} & 60.7 & 60.3 \\
\midrule
Baseline 1  & 2765 & 5  & 59.2 & 59.6 & 55.1 & 57.3 & 58.3 \\
Baseline 2  & 2765 & -- & 51.2 & 44.4 & 53.3 & 48.9 & 48.4 \\
\bottomrule
\end{tabular}
\vspace{-10pt}
\end{table}

\subsection{Approach Generalization}
\label{section_generalization}
 To assess the generalizability of our modeling pipeline and benchmark it against existing methods, we conducted external validation using the TILES-18 dataset \cite{Mundnich_Booth_L_et_al._2020}. We compared our approach to a prior study \cite{Pranjal_Seshadri_etal_2023} that predicted state anxiety using features derived from Fitbit and other devices. Since our model relies on watch-sensed heart rate (HR), we implemented two baseline versions: one using all Fitbit-derived features (e.g., cardio, fat burn, steps, sleep) and another using only HR features, as in the original study. As shown in Table \ref{tab:comparison_generalization}, our meta-learner model substantially outperformed both baselines—for instance, achieving a 7.9\% higher balanced accuracy than the all-feature model.

\begin{table}[htbp]

\caption{TILES-18 evaluation. F = Features.}
\label{tab:comparison_generalization}
\small
\setlength{\tabcolsep}{4pt}
\begin{tabular}{lccccccc}
\toprule
Model & EMAs & F. & Prec. & Rec. & Spec. & BA & F1 \\
\midrule
Meta-learner (Ours) & 10095 & 7 & \textbf{61.4} & \textbf{50.9} & \textbf{67.4} & \textbf{59.1} & \textbf{61.3} \\
HR-only \cite{Pranjal_Seshadri_etal_2023} & 10095 & 1 & 51.3 & 3.6 & 95.7 & 49.7 & 49.1 \\
All Fitbit \cite{Pranjal_Seshadri_etal_2023} & 10095 & 25 & 54.4 & 23.7 & 78.8 & 51.2 & 54.7 \\
\bottomrule
\end{tabular}
\vspace{-10pt}
\end{table}

\subsection{Ablation Study}
\label{ablation_study}

 To evaluate the contributions of the transfer learning (TL) model and the meta-learner, we conducted an ablation study. First, we assessed the TL model alone—without the meta-learner—on both our dataset and the external TILES-18 dataset. On our dataset, the TL model achieved 58.5\% recall and 36.7\% specificity. On TILES-18, it achieved 38.5\% recall and 58.9\% specificity. Despite lower overall performance, the TL model outperformed previously published approaches \cite{Pranjal_Seshadri_etal_2023} in recall when evaluated on 10,095 EMA responses from TILES-18. As shown in Table \ref{tab:comparison_generalization}, recall for the prior models ranged from just 3.6\% (HR-only features) to 23.7\% (all Fitbit features), compared to 38.5\% with our TL-only model and 50.9\% with meta-learner.

We trained a trait-only model using four selected trait measures, applying the same feature selection, classifier (Logit), and hyperparameters as the meta-learner. While overall metrics were comparable (e.g., BA: 60.4\% vs. 60.16\%; F1: 60.4\% vs. 59.2\%), the trait-only model failed to capture intra- and inter-day variability, consistently predicting the same class per participant (BA: 0\%, 50\%, or 100\%). In contrast, our meta-learner produced more temporally sensitive, participant-specific predictions. For example, in our dataset, 7 participants had per-participant BA between 50\%–100\% and 9 between 0\%–50\%, with similar results on the TILES-18 dataset. These differences reflect the static nature of trait-only inputs versus the temporal variation in TL-derived probabilities used by the meta-learner.

\vspace{-3pt}
\section{Conclusion and Future Work}
\vspace{-3pt}
 We propose a model that leverages watch-sensed data to predict state anxiety. While it outperforms baseline models, there is still considerable room for improvement to support more precise and personalized interventions. Future work will explore incorporating additional sensor modalities and enabling on-device detection of timely intervention opportunities. In parallel, strategies such as thresholds jointly determined by clinicians and users (e.g., lowering the threshold to increase sensitivity) could help balance sensitivity and specificity, thereby mitigating potential negative effects. Also, our deep learning–based approach is limited by interpretability, which future work could address.


\printbibliography

@article{toner_wearable_2025,
	title = {Wearable {Sensor}-based {Multimodal} {Physiological} {Responses} of {Socially} {Anxious} {Individuals} in {Social} {Contexts} on {Zoom}},
	issn = {1949-3045},
	url = {https://ieeexplore.ieee.org/abstract/document/10971224},
	doi = {10.1109/TAFFC.2025.3562787},
	journal = {IEEE Transactions on Affective Computing},
	author = {Toner, Emma R. and Rucker, Mark and Wang, Zhiyuan and Larrazabal, Maria A. and Cai, Lihua and Datta, Debajyoti and Lone, Haroon and Boukhechba, Mehdi and Teachman, Bethany A. and Barnes, Laura E.},
	year = {2025},
	pages = {1--12},
}

@article{kessler_twelvemonth_2012,
	title = {Twelve‐month and lifetime prevalence and lifetime morbid risk of anxiety and mood disorders in the {United} {States}},
	volume = {21},
	journal = {International journal of MPR},
	author = {Kessler, Ronald C. and Petukhova, Maria and Sampson, Nancy A. and Zaslavsky, Alan M. and Wittchen, Hans-Ullrich},
	year = {2012},
	note = {ISBN: 1049-8931
Publisher: Wiley Online Library},
	pages = {169--184},
}

@article{jacobson_digital_2022,
	title = {Digital biomarkers of anxiety disorder symptom changes: {Personalized} deep learning models using smartphone sensors accurately predict anxiety symptoms from ecological momentary assessments},
	volume = {149},
	issn = {0005-7967},
	shorttitle = {Digital biomarkers of anxiety disorder symptom changes},
	url = {https://www.sciencedirect.com/science/article/pii/S0005796721002126},
	doi = {10.1016/j.brat.2021.104013},
	journal = {Behaviour Research and Therapy},
	author = {Jacobson, Nicholas C. and Bhattacharya, Sukanya},
	month = feb,
	year = {2022},
	pages = {104013},
}

@article{boukhechba_predicting_2018,
	title = {Predicting {Social} {Anxiety} {From} {Global} {Positioning} {System} {Traces} of {College} {Students}: {Feasibility} {Study}},
	volume = {5},
	copyright = {Unless stated otherwise, all articles are open-access distributed under the terms of the Creative Commons Attribution License (http://creativecommons.org/licenses/by/2.0/), which permits unrestricted use, distribution, and reproduction in any medium, provided the original work ("first published in the Journal of Medical Internet Research...") is properly cited with original URL and bibliographic citation information. The complete bibliographic information, a link to the original publication on http://www.jmir.org/, as well as this copyright and license information must be included.},
	shorttitle = {Predicting {Social} {Anxiety} {From} {Global} {Positioning} {System} {Traces} of {College} {Students}},
	url = {https://mental.jmir.org/2018/3/e10101},
	doi = {10.2196/10101},
	
	number = {3},
	
	journal = {JMIR Mental Health},
	author = {Boukhechba, Mehdi and Chow, Philip and Fua, Karl and Teachman, Bethany A. and Barnes, Laura E.},
	month = jul,
	year = {2018},
	pages = {e10101},
}

@article{rashid_predicting_2020,
	title = {Predicting {Subjective} {Measures} of {Social} {Anxiety} from {Sparsely} {Collected} {Mobile} {Sensor} {Data}},
	volume = {4},
	url = {https://dl.acm.org/doi/10.1145/3411823},
	doi = {10.1145/3411823},
	journal = {ACM IMWUT},
	author = {Rashid, Haroon and Mendu, Sanjana and Daniel, Katharine E. and Beltzer, Miranda L. and Teachman, Bethany A. and Boukhechba, Mehdi and Barnes, Laura E.},
	year = {2020},
	pages = {109:1--109:24},
}

@article{sahu_wearable_2024,
	title = {Wearable {Technology} {Insights}: {Unveiling} {Physiological} {Responses} {During} {Three} {Different} {Socially} {Anxious} {Activities}},
	volume = {2},
	shorttitle = {Wearable {Technology} {Insights}},
	url = {https://doi.org/10.1145/3663671},
	doi = {10.1145/3663671},
	number = {2},
	journal = {ACM JCSS},
	author = {Sahu, Nilesh Kumar and Gupta, Snehil and Lone, Haroon},
	month = jun,
	year = {2024},
	pages = {27:1--27:23},
}

@article{caspi_moving_1988,
	title = {Moving away from the world: {Life}-course patterns of shy children},
	volume = {24},
	issn = {1939-0599},
	shorttitle = {Moving away from the world},
	doi = {10.1037/0012-1649.24.6.824},
	number = {6},
	journal = {Developmental Psychology},
	author = {Caspi, Avshalom and Elder Jr., Glen H. and Bem, Daryl J.},
	year = {1988},
	note = {Place: US
Publisher: American Psychological Association},
	pages = {824--831},
}

@article{Mundnich_Booth_L_et_al._2020, title={TILES-2018, a longitudinal physiologic and behavioral data set of hospital workers}, volume={7}, url={http://dx.doi.org/10.1038/s41597-020-00655-3}, DOI={10.1038/s41597-020-00655-3}, number={1}, journal={Scientific Data}, publisher={Springer Science and Business Media LLC}, author={Mundnich, Karel and Booth, Brandon M. and L’Hommedieu, Michelle and Feng, Tiantian and Girault, Benjamin and L’Hommedieu, Justin and Wildman, Mackenzie and Skaaden, Sophia and Nadarajan, Amrutha and Villatte, Jennifer L. and Falk, Tiago H. and Lerman, Kristina and Ferrara, Emilio and Narayanan, Shrikanth}, year={2020}}

@article{KINDRED2022102640,
title = {Long-term outcomes of cognitive behavioural therapy for social anxiety disorder: A meta-analysis of randomised controlled trials},
journal = {Journal of Anxiety Disorders},
volume = {92},
pages = {102640},
year = {2022},
issn = {0887-6185},
doi = {https://doi.org/10.1016/j.janxdis.2022.102640},
url = {https://www.sciencedirect.com/science/article/pii/S088761852200113X},
author = {Reuben Kindred and Glen W. Bates and Nicholas L. McBride}
}

@article{Larrazabal_Wang_Rucker_Toner_Boukhechba_Teachman_Barnes_2025, title={Understanding State Social Anxiety in Virtual Social Interactions using Multimodal Wearable Sensing Indicators}, url={https://arxiv.org/abs/2503.15637}, DOI={10.48550/ARXIV.2503.15637}, journal={IEEE SMARTCOMP'25 [In Press]}, author={Larrazabal, Maria A. and Wang, Zhiyuan and Rucker, Mark and Toner, Emma R. and Boukhechba, Mehdi and Teachman, Bethany A. and Barnes, Laura E.}, year={2025} }

@article{Prakash_Madanmohan_2005, title={How to Tell Heart Rate From an ECG? (Learning Objects \#769 and \#878)}, volume={29}, url={http://dx.doi.org/10.1152/advan.00013.2005}, DOI={10.1152/advan.00013.2005}, number={2}, journal={Advances in Physiology Education}, publisher={American Physiological Society}, author={Prakash, E. S. and Madanmohan}, year={2005}, month=june, pages={57–57},  }

@InProceedings{He_2016_CVPR,
author = {He, Kaiming and Zhang, Xiangyu and Ren, Shaoqing and Sun, Jian},
title = {Deep Residual Learning for Image Recognition},
booktitle = {IEEE CVPR},
month = {June},
year = {2016}
}

@article{Ebbehoj_Thunbo_Andersen_Glindtvad_Hulman_2022, title={Transfer learning for non-image data in clinical research: A scoping review}, volume={1}, url={http://dx.doi.org/10.1371/journal.pdig.0000014}, DOI={10.1371/journal.pdig.0000014}, number={2}, journal={PLOS Digital Health}, publisher={Public Library of Science (PLoS)}, author={Ebbehoj, Andreas and Thunbo, Mette Østergaard and Andersen, Ole Emil and Glindtvad, Michala Vilstrup and Hulman, Adam}, year={2022}, month=feb, pages={e0000014},  }

@article{Xiong_Huang_Barnes_Gerber_2016, title={Sensus: a cross-platform, general-purpose system for mobile crowdsensing in human-subject studies}, url={http://dx.doi.org/10.1145/2971648.2971711}, DOI={10.1145/2971648.2971711}, journal={ACM UbiComp'16}, publisher={ACM}, author={Xiong, Haoyi and Huang, Yu and Barnes, Laura E. and Gerber, Matthew S.}, year={2016}}

@INPROCEEDINGS{Shahmohammadi2017-hm,
  title           = "Smartwatch based activity recognition using active
                     learning",
  booktitle       = "IEEE/ACM CHASE'17",
  author          = "Shahmohammadi, Farhad and Hosseini, Anahita and King,
                     Christine E and Sarrafzadeh, Majid",
  publisher       = "IEEE",
  pages           = "321--329",
  year            =  2017
}

@article{Pranjal_Seshadri_etal_2023, title={Toward Privacy-Enhancing Ambulatory-Based Well-Being Monitoring: Investigating User Re-Identification Risk in Multimodal Data}, url={http://dx.doi.org/10.1109/ICASSP49357.2023.10096235}, DOI={10.1109/icassp49357.2023.10096235}, journal={IEEE ICASSP'23}, publisher={IEEE}, author={Pranjal, Ravi and Seshadri, Ranjana and Kumar Sanath Kumar Kadaba, Rakesh and Feng, Tiantian and Narayanan, Shrikanth S. and Chaspari, Theodora}, year={2023}, month=june, pages={1–5} }

@article{Makowski_NeuroKit2_2021, title={NeuroKit2: A Python toolbox for neurophysiological signal processing}, volume={53}, url={http://dx.doi.org/10.3758/s13428-020-01516-y}, DOI={10.3758/s13428-020-01516-y}, number={4}, journal={Behavior Research Methods}, publisher={Springer Science and Business Media LLC}, author={Makowski, Dominique and Pham, Tam and Lau, Zen J. and Brammer, Jan C. and Lespinasse, François and Pham, Hung and Schölzel, Christopher and Chen, S. H. Annabel}, year={2021}, month=feb, pages={1689–1696}, language={en} }

\end{document}